\documentclass{article} 
\usepackage{iclr2027_conference,times}

\usepackage{amsmath,amsfonts,bm}

\def\eqref#1{equation~\ref{#1}}

\def\1{\bm{1}}

\DeclareMathAlphabet{\mathsfit}{\encodingdefault}{\sfdefault}{m}{sl}
\SetMathAlphabet{\mathsfit}{bold}{\encodingdefault}{\sfdefault}{bx}{n}

\usepackage{capt-of}
\usepackage[T1]{fontenc}
\usepackage{amsmath,amssymb,amsthm}
\usepackage{graphicx,booktabs,array,multirow}
\usepackage{xspace}
\usepackage{algorithm,algpseudocode}
\usepackage[table]{xcolor}
\usepackage{longtable}
\usepackage{hyperref}
\usepackage{url}
\usepackage{wrapfig}
\newcommand{\result}[2]{%
  \ensuremath{#1\,\mathord{\raisebox{0.12ex}{\scalebox{0.75}{$\pm#2$}}}}%
}

\newcommand{\bestresult}[2]{%
  \ensuremath{\mathbf{#1}\,\mathord{\raisebox{0.12ex}{\scalebox{0.75}{$\boldsymbol{\pm}\mathbf{#2}$}}}}%
}

\newcommand{\secondresult}[2]{%
  \underline{\ensuremath{#1\,\mathord{\raisebox{0.12ex}{\scalebox{0.75}{$\pm#2$}}}}}%
}

\newcommand{\method}{SERO\xspace}

\title{SERO: Learning from Executable Oracle Continuations for Long-Horizon Agentic Reinforcement Learning}

\author{
\textbf{Dingyi Rong$^{1,2}$, Yue Shi$^{1,\dagger}$, Chaofan Ma$^{2,\dagger}$, Zongrui Wang$^{1}$, Jiezhang Cao$^{2}$, Yixin Cao$^{2}$,}\\
\textbf{Liang He$^{3}$, Ning Liu$^{2,\dagger}$}\\[4pt]
\normalfont
$^{1}$Shanghai Artificial Intelligence Laboratory\\
$^{2}$Shanghai Jiao Tong University\\
$^{3}$Tongji University\\
$^{\dagger}$Corresponding authors
}

\iclrfinalcopy 
\begin{document}
\maketitle
\fancyhead{}
\newcommand{\yue}[1]{\textcolor{red}{\textbf{yue:} #1}}

\begin{abstract}
Long-horizon agentic RL is constrained by the futures visited by the current policy. When rollouts fail to reach useful behaviors, improved credit assignment cannot provide supervision for states and actions that were never encountered. On-policy distillation (OPD) introduces teacher guidance on policy-induced states, and recent multi-turn variants improve how this guidance is scheduled, replayed, refined, or validated. However, they do not directly use the oracle-controlled future itself as the training trajectory. We introduce \textbf{State-Rooted Executable Rollout Optimization (SERO)}, which augments on-policy agentic RL with executable oracle continuations. SERO restores a state actually reached by the current policy and lets a training-only oracle continue interacting with the real environment, making the downstream observations and oracle actions themselves available for learning. It then applies \textbf{policy-reachability weighting}, which weights each oracle step by the frozen behavior policy's compatibility with the preceding oracle actions. This allows deeper supervision to become effective as the policy becomes more compatible with the oracle branch. The oracle loss is optimized separately from the base RL estimator and no oracle is used at inference time. We evaluate SERO on two challenging interactive-agent benchmarks, ALFWorld and WebShop, using Qwen2.5-1.5B-Instruct and Qwen2.5-7B-Instruct. On ALFWorld, learning the full executable continuation is substantially more effective than supervising only the root oracle action, while policy-reachability weighting further improves out-of-distribution performance.
\end{abstract}

\section{Introduction}
Large language models and vision-language models are increasingly used as the decision-making backbones of interactive agents for embodied control, web navigation, tool use, and software tasks~\citep{yao2023react,shinn2023reflexion,schick2023toolformer,zitkovich2023rt2,wang2024voyager,xie2024osworld,zhang2026r3dp,hong2024cogagent,gur2024webagent,xia2026roboprocessbench,gou2025universalgrounding,wei2025swerl,rong2026h2rbench}. 
Unlike single-turn reasoning, an agent must repeatedly observe the environment, choose an action, receive a new observation, and continue until the task is finished~\citep{ma2026rere}. 
This sequential decision-making setting provides a natural opportunity for reinforcement learning (RL), where policies can be improved directly from environment feedback~\citep{sutton2018rl,schulman2017ppo,ouyang2022instructgpt,guo2025deepseekr1,tan2024practice,zhai2024vlmagent,bai2024digirl}. 
In particular, critic-free group-based methods such as RLOO~\citep{ahmadian2024rloo}, GRPO~\citep{shao2024deepseekmath}, and their recent variants are attractive for large-scale agent training, because they avoid training a separate value model and reduce the overhead associated with critic-based optimization~\citep{yu2025dapo,liu2025drgrpo,lin2025cppo,zheng2025gspo}.

However, these methods typically assign a single outcome-based advantage to all actions in a trajectory, which makes credit assignment a central difficulty as interaction horizons grow. 
An episode may contain many decisions, while reward is often sparse or delayed~\citep{peng2019awr,wen2024actiondecomp,chen2025longhorizon,song2024trialerror,wang2025ragen,jin2025searchr1}.
Recent work therefore seeks more precise feedback for the trajectories collected by the current policy~\citep{feng2025gigpo,he2026hgpo,cheng2026graphgpo}. 
These methods substantially improve how sampled experience is credited. 
Nevertheless, the learning signal is still bounded by what appears in the rollout group. 
If every rollout misses a useful action sequence, finer attribution can rank the sampled behaviors more accurately, but it cannot provide the missing behavior as a positive target. 
This limitation is particularly important early in training and on difficult tasks, where successful branches may be absent from an entire group.

\begin{figure}[t]
\centering
\includegraphics[width=0.98\linewidth]{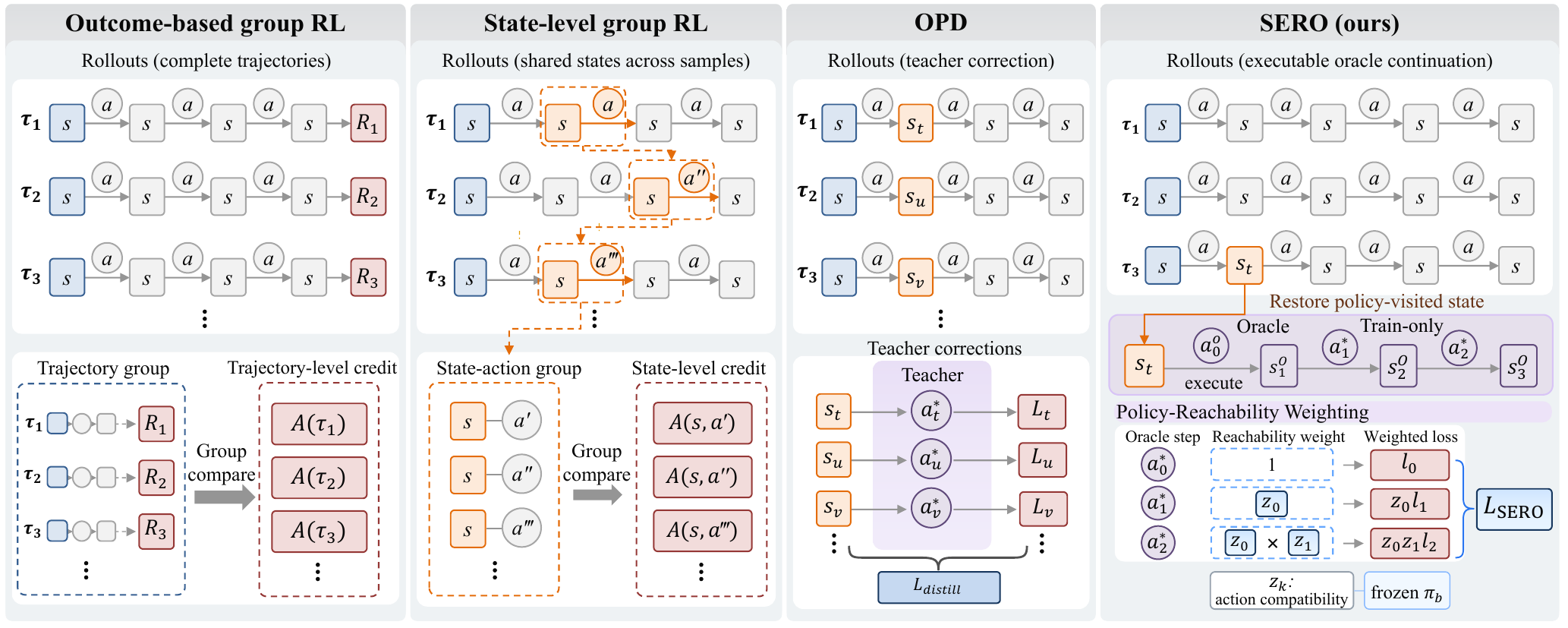}
\caption{Comparison of training paradigms and learning signals. Outcome-based group RL assigns trajectory-level credit, state-level group RL compares sampled actions at shared states, and OPD distills a single corrective action.
In contrast, \method restores a state reached by the current policy, executes a multi-step oracle continuation, and applies policy-reachability weighting to downstream action losses.}
\label{fig:comparison}
\end{figure}

In such cases, the missing targets must be supplied from outside the rollout group, which motivates teacher guidance.
Interactive imitation and off-policy guidance learn from expert behavior on learner-induced inputs~\citep{ross2011dagger,sun2017aggrevated,yan2025luffy}. 
OraRL shows that privileged supervision can be optimized directly, while on-policy distillation (OPD) queries a teacher on states reached by the current policy~\citep{li2026orarl,agarwal2024gkd,sun2026gapd}. 
Recent multi-turn OPD methods extend teacher guidance over multiple steps or use subsequent policy rollouts to validate an intervention~\citep{wang2026tcod,liao2026reopd,chen2026ftb}. 
However, their supervision is still centered on a corrective action, a teacher prefix, or the outcome of the intervention, rather than the full missing branch.
In interactive environments, executing a correction changes the next observation and therefore the decisions that follow~\citep{ross2011dagger,ma2021bsiris,liu2022tis}. 
When these downstream transitions are not directly learned, the training signal captures only part of the benefit of the correction and leaves its long-horizon consequences under-supervised.

We propose \textbf{State-Rooted Executable Rollout Optimization (SERO)} to learn from executable multi-step oracle continuations rooted at policy-visited states. As shown in Figure \ref{fig:comparison}, SERO first collects the ordinary on-policy trajectories required by the base RL algorithm. It then selects a state actually reached by the current policy, restores the environment at that state, and lets a training-only oracle continue interacting for multiple steps. Every oracle action is executed in the environment, so each later target is conditioned on the real observation produced by the preceding oracle actions. The resulting continuation therefore shows not only what action should be taken at the root state, but also the sequence of decisions that follows once that correction is actually carried out.

Deeper states in an oracle continuation are increasingly off-policy, since reaching them requires the current policy to follow the preceding oracle actions. We therefore introduce \textbf{policy-reachability weighting}. A frozen behavior policy scores each oracle action, and the supervision weight at step $k$ is determined by the compatibility of the preceding oracle prefix under the behavior policy. Within a branch admitted by the improvement gate, the root target receives unit prefix weight, while deeper targets receive larger weights as the policy becomes more compatible with the oracle prefix. This down-weights supervision on poorly supported oracle prefixes and allows the effective supervision horizon to extend as compatibility improves. The oracle objective is added as an auxiliary loss, leaving the rewards, advantages, and policy ratios of the base RL algorithm unchanged, and the oracle is used only during training.

We evaluate SERO on ALFWorld~\citep{shridhar2021alfworld} and WebShop~\citep{yao2022webshop} with Qwen2.5-1.5B-Instruct and Qwen2.5-7B-Instruct. Adding SERO to GraphGPO or HGPO improves ALFWorld success and WebShop task success at both scales. We also compare against prompting and strong RL baselines, and ablate root-only oracle targets, uniformly weighted full oracle continuations, and policy-reachability weighting. On ALFWorld, full executable continuations outperform root corrections, with reachability weighting adding further gains, especially on unseen tasks.


Our contributions are summarized as follows:
\begin{itemize}
    \item We propose \textbf{State-Rooted Executable Rollout Optimization (SERO)}, which turns an oracle correction into a multi-step source of executable training experience. Starting from a state reached by the current policy, SERO executes the oracle in the environment and learns the downstream oracle actions under the observations created by that intervention.
    \item We introduce \textbf{policy-reachability weighting} as a reachability-aware way to learn from oracle-induced states. By tying each downstream target to the behavior-policy compatibility of the preceding oracle prefix, SERO suppresses supervision on poorly supported oracle-induced states and lets deeper guidance become effective as the policy improves.
    \item We design SERO as an auxiliary objective that complements existing on-policy credit assignment while leaving the base RL credit estimator unchanged. Experiments at both model scales show improvements in ALFWorld success and WebShop strict success for both GraphGPO and HGPO, while controlled ablations isolate the effects of executable continuation learning and reachability weighting.
\end{itemize}

\section{Related Work}

\paragraph{Group-based RL for long-horizon agents.}
Group-based RL has become a practical alternative to actor--critic training for large language models by replacing a learned value function with relative comparisons among sampled rollouts~\citep{ahmadian2024rloo,shao2024deepseekmath,yu2025dapo,liu2025drgrpo,lin2025cppo,zheng2025gspo}. In interactive tasks, however, sparse and delayed rewards make it difficult to determine which decisions actually helped. Recent work has also extended online RL to self-evolving web curricula and end-to-end multi-turn interaction, while newer methods study hindsight credit assignment and uncertainty-guided exploration in long-horizon agents
\citep{qi2025webrl,wei2025webagent,tan2026hindsight,wang2026t}. Recent agentic RL therefore moves from one advantage per trajectory toward finer signals recovered from the collected experience, using repeated states, interaction history, or cross-trajectory transition structure to refine credit assignment~\citep{zhou2024archer,luo2025agentlightning,feng2025gigpo,he2026hgpo,cheng2026graphgpo,wang2026steppo}. These methods improve how the policy learns from its own rollouts, but they share a structural limit: supervision is still extracted from behavior already present in the sampled group. If every rollout misses a useful branch, a better estimator can re-rank the observed decisions but cannot provide that missing branch as a positive target.

\paragraph{Teacher guidance on learner-induced states.}
Imitation and distillation address the same difficulty from another direction by querying an expert on states induced by the learner rather than only replaying fixed demonstrations~\citep{ross2011dagger,sun2017aggrevated,agarwal2024gkd,gu2024minillm,yan2025luffy}. Related approaches also construct corrective supervision from failed trajectories or expert interaction experience
\citep{yuan2025agent,lu2026experience}. OPD and recent multi-turn extensions bring this principle to language-model post-training, aligning teacher supervision with the current policy and extending it through prefixes, refined trajectories, or short teacher interventions~\citep{li2026videoopd,sun2026gapd,wang2026tcod,liao2026reopd,jiang2026trd,chen2026ftb}. Their common goal is to align teacher guidance with the learner's current behavior. SERO instead learns the executable future induced by the correction: after restoring a policy-visited state, the oracle continues the branch, and the resulting observations and downstream actions are used directly for training.

\paragraph{Oracle-guided optimization.}
Privileged information can also enter RL directly as an optimization signal. OraRL shows that an annotation can serve as a positive target rather than only a reward reference, while CAST uses solver information to refine turn-level credit~\citep{li2026orarl,wang2026cast}. SERO shares the motivation that finite on-policy samples can benefit from reliable privileged supervision, but interactive dynamics make the target fundamentally sequential: executing an oracle action changes the environment and creates new states. SERO therefore keeps the original on-policy estimator unchanged and learns a separate executable oracle continuation rooted at on-policy experience. Policy-reachability weighting then controls the influence of deeper oracle-induced states according to the behavior-policy compatibility of the preceding oracle prefix.

\section{Preliminaries}
\label{sec:prelim}

\subsection{Long-Horizon Agentic Reinforcement Learning}
We consider the standard sequential decision-making setting~\citep{sutton2018rl} used in long-horizon agentic RL~\citep{feng2025gigpo,he2026hgpo,cheng2026graphgpo}. At step $t$, the policy observes an agent state $s_t\in\mathcal S$, which contains the current environment observation and the interaction context exposed by the agent interface, and samples an action $a_t\sim\pi_\theta(\cdot\mid s_t)$. The environment executes $a_t$ and returns reward $r_t$ and the next state $s_{t+1}$. A trajectory is $\tau=\{(s_t,a_t,r_t)\}_{t=0}^{T-1}$ with return $R(\tau)$. In long-horizon tasks such as ALFWorld~\citep{shridhar2021alfworld} and WebShop~\citep{yao2022webshop}, rewards are sparse or delayed, so the contribution of an individual action is difficult to infer from the final outcome alone~\citep{wang2026tracebench}. During training, we additionally save the simulator snapshot corresponding to each visited state $s_t$ so that the same decision point can be restored for oracle continuation. This snapshot is used only for training and is never included in the policy input.

\subsection{Group-Based Policy Optimization and Oracle Supervision}
\label{sec:base_rl}
Group-based RL replaces an explicit learned critic with statistics computed from sampled groups~\citep{ahmadian2024rloo,shao2024deepseekmath}. Following clipped policy optimization~\citep{schulman2015trpo,schulman2017ppo}, it samples $N$ trajectories for the same task using a frozen behavior policy $\pi_b=\pi_{\theta_{\rm old}}$. For example, GRPO assigns
\begin{equation}
    \widehat A_i^{\rm GRPO}
    =\frac{R(\tau_i)-\mu_R}{\sigma_R+\epsilon_{\rm num}},
    \qquad
    \mu_R=\frac{1}{N}\sum_{j=1}^{N}R(\tau_j),
\label{eq:grpo_adv}
\end{equation}
where $\sigma_R$ is the group reward standard deviation and $\epsilon_{\rm num}>0$ is a small constant for numerical stability. Recent agentic variants provide finer credit by comparing repeated states, conditioning on history, or using the transition structure of sampled 
trajectories~\citep{feng2025gigpo,he2026hgpo,cheng2026graphgpo}. We denote the advantage supplied by the chosen base method as $\widehat A^{\rm base}_{i,t}$ and schematically write its clipped objective as
\begin{equation}
    \mathcal L_{\rm base}
    =-\frac{1}{M}\sum_{i,t}
    \min\!\left(\rho_{i,t}\widehat A^{\rm base}_{i,t},
    \bar\rho_{i,t}\widehat A^{\rm base}_{i,t}\right)
    +\beta\mathcal L_{\rm reg},
\label{eq:base_loss}
\end{equation}
where $T_i$ is the number of valid decision steps in trajectory $i$, $M=\sum_i T_i$ is the number of valid decision steps in the batch, $\rho_{i,t}=\pi_\theta(a_{i,t}\mid s_{i,t})/\pi_b(a_{i,t}\mid s_{i,t})$, and $\bar\rho_{i,t}=\operatorname{clip}(\rho_{i,t},1-\epsilon_c,1+\epsilon_c)$ for clipping threshold $\epsilon_c>0$. The optional term $\beta\mathcal L_{\rm reg}$, with $\beta\geq0$, denotes any regularization used by the base learner (e.g., a KL penalty). 


These methods assign credit using trajectories generated by the policy itself. A different source of supervision is a training-time oracle. At a policy-visited state $s_t$, the oracle can provide a corrective action
$a_t^\star=\mathcal O(s_t)$.
In an interactive environment, this correction changes the next observation and the rest of the trajectory. Training only on the first oracle action therefore ignores the useful decisions that follow. Our method uses the resulting oracle continuation as multi-step supervision.

\section{Proposed Method}
\label{sec:method}

\subsection{Overview}

\begin{figure}[t]
\centering
\includegraphics[width=0.98\linewidth]{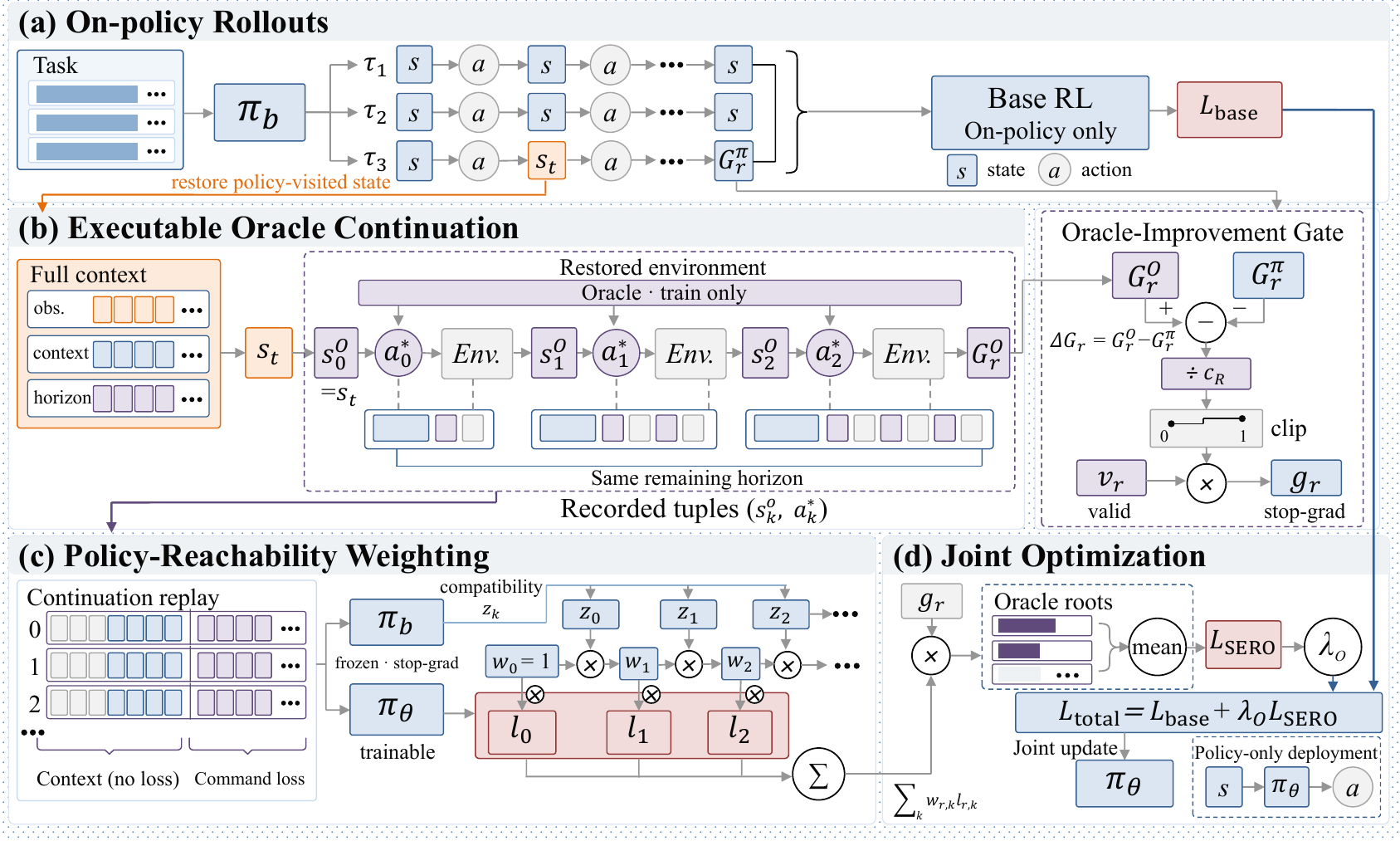}
\caption{Overview of \method.
(a) The behavior policy collects on-policy rollouts for the base RL objective.
(b) From a selected policy-visited state, \method restores the environment, executes an oracle continuation under the same remaining horizon, and computes a detached improvement gate.
(c) The frozen behavior policy provides compatibility scores used to form policy-reachability weights for downstream oracle-action losses.
(d) The root-averaged auxiliary loss is combined with the base RL loss in one policy update. The oracle and environment restoration are used only during training, and only the learned policy is used at inference. }
\label{fig:overview}
\end{figure}

SERO is a modular, oracle-augmentation framework that can be integrated with different on-policy RL objectives without modifying their underlying credit-assignment mechanisms. 
As is shown in Figure \ref{fig:overview}, each training iteration has three stages. 
First, the base RL method collects ordinary on-policy rollouts and computes its original RL objectives. 
Second, SERO selects a state from an on-policy trajectory, restores the environment to that decision point, and lets an oracle execute a fresh multi-step continuation. 
Third, the oracle actions are optimized with weights that reflect policy support for the oracle prefix leading to each downstream state. 
Oracle data are used only by the auxiliary branch and never enter the reward statistics, advantages, or importance ratios of the base method.


\subsection{Executable Oracle Continuation}
\label{sec:oracle_continuation}
To obtain downstream oracle targets that remain valid after an intervention, SERO constructs a fresh oracle continuation from a state reached by the current policy. 
Let $s_t$ be a selected pre-action state from an on-policy rollout. 
We restore the corresponding simulator snapshot in a separate environment copy, preserving the same decision point and remaining horizon. 
The restored state initializes the oracle branch as $s_0^O=s_t$, where $s_k^O$ denotes the state at oracle step $k$.
\begin{equation}
    a_k^{\star}=\mathcal O(s_k^O),\qquad
    (s_{k+1}^O,r_k^O)=\mathcal E(s_k^O,a_k^{\star}),
\label{eq:oracle_transition}
\end{equation}
where $\mathcal E$ denotes one environment transition. Repeating this process until task completion or exhaustion of the remaining interaction budget gives
\begin{equation}
    \tau_r^O=
    \big\{(s_k^O,a_k^\star,r_k^O)\big\}_{k=0}^{L_r-1},
\label{eq:branch}
\end{equation}
where $L_r$ is the realized continuation length. 
The oracle continuation is generated step by step from the states actually reached in the restored environment. 
The oracle may use privileged training information for action selection, while the trainable policy is optimized only from the policy-visible state $s_k^O$. 
The on-policy suffix and oracle continuation share the same restored root and remaining horizon, allowing their returns to be compared from the same decision point.

\subsection{Policy-Reachability Weighting}
\label{sec:weighting}

Executing the oracle makes each downstream target valid in the environment, but does not make it equally relevant to the current policy.
The restored root $s_0^O$ comes directly from an on-policy trajectory, whereas reaching a deeper state $s_k^O$ requires the current policy to follow the preceding oracle actions. 
SERO captures this with a prefix-reachability weight.

Let the oracle command at step $k$ be
$a_k^\star=(y_{k,1}^\star,\ldots,y_{k,m_k}^\star)$.
Given the frozen behavior policy $\pi_b$, we measure the local compatibility of this command by its length-normalized likelihood:
\begin{equation}
z_k=
\exp\!\left(
\frac{1}{m_k}
\sum_{j=1}^{m_k}
\log \pi_b
\left(
y_{k,j}^\star
\mid
s_k^O,c_k,y_{k,<j}^\star
\right)
\right),
\label{eq:step_compat_command}
\end{equation}
where $c_k$ is the reasoning prefix produced by $\pi_b$ at $s_k^O$ and used as replay context; $z_k$ is therefore a length-normalized compatibility score rather than a marginal action probability.

The weight for the $k$-th oracle target is then
\begin{equation}
w_0=1,\qquad
w_k=\prod_{j=0}^{k-1}z_j,
\qquad
w_{k+1}=w_k z_k.
\label{eq:prefix}
\end{equation}
Thus, $w_k$ measures how well the current policy supports the oracle prefix leading to the $k$-th target. The root action always receives unit prefix weight, while deeper supervision is reduced when earlier oracle actions have low compatibility under the behavior policy. As the policy becomes more compatible with the oracle prefix, the weights decay more slowly and supervision can extend further along the continuation. The corresponding discrete-action formulation is given in Appendix~\ref{app:interpretation}.

\subsection{Oracle Continuation Objective}

Reachability determines how strongly each oracle target is supervised, but not whether the oracle continuation improves upon the policy’s own future from the same root. SERO therefore uses a branch-level improvement gate to control its contribution to the auxiliary objective.

Index attempted oracle roots by $r=1,\ldots,N_{\rm root}$. 
Let $G_r^{\pi}$ and $G_r^{O}$ denote the realized returns of the on-policy suffix and oracle continuation, respectively, over the same remaining horizon.
We define
\begin{equation}
    \Delta G_r=G_r^{O}-G_r^{\pi},
    \qquad
    g_r=v_r\,\operatorname{clip}\!\left(\frac{\Delta G_r}{c_R},0,1\right),
\label{eq:gate}
\end{equation}
where $v_r\in\{0,1\}$ indicates a valid restored branch and $c_R>0$ is a reward-normalization scale, with $c_R=1$ for binary success returns. 

For a contributing oracle branch, SERO trains the policy to reproduce each oracle action using a supervised likelihood loss.
For a language command,
\begin{equation}
    \ell_k(\theta)
    =-\frac{1}{m_k}\sum_{j=1}^{m_k}
      \log\pi_\theta(y_{k,j}^{\star}\mid s_k^O,c_k,y_{k,<j}^{\star}),
\label{eq:step_loss_command}
\end{equation}
which reduces to $-\log\pi_\theta(a_k^\star\mid s_k^O)$ for a categorical action.

Let $L_r$ be the continuation length of root $r$, $w_{r,k}$ and $\ell_{r,k}$ its step-$k$ reachability weight and oracle-action loss, and $\operatorname{sg}[\cdot]$ the stop-gradient operator. Combining branch-level improvement, prefix reachability, and oracle supervision gives
\begin{equation}
\mathcal L_{\mathrm{SERO}}(\theta)
=\frac{1}{\max(N_{\rm root},1)}
\sum_{r=1}^{N_{\rm root}}
\underbrace{\operatorname{sg}[g_r]}_{\text{branch improvement}}
\sum_{k=0}^{L_r-1}
\underbrace{\operatorname{sg}[w_{r,k}]}_{\text{prefix reachability}}
\cdot
\underbrace{\ell_{r,k}(\theta)}_{\text{oracle supervision}}.
\label{eq:loss}
\end{equation}
The local compatibility score $z_k$ enters through the construction of $w_{r,k}$. Both branch and reachability weights are computed from pre-update quantities and detached from the gradient.

SERO thus adds a decoupled oracle-supervision objective alongside the base RL loss, without altering the on-policy advantages or requiring importance-ratio correction.
\begin{equation}
    \mathcal L_{\rm total}(\theta)
    =\mathcal L_{\rm base}(\theta)
     +\lambda_O\mathcal L_{\mathrm{SERO}}(\theta),
\label{eq:total}
\end{equation}
where $\lambda_O\geq0$ controls the strength of oracle-continuation supervision. On-policy trajectories alone determine the base learner's rewards, group statistics, advantages, and policy ratios. This makes SERO compatible with different credit estimators, including GRPO \citep{shao2024deepseekmath}, GiGPO \citep{feng2025gigpo}, HGPO \citep{he2026hgpo}, and GraphGPO \citep{cheng2026graphgpo}. The oracle and environment-restoration machinery are used only during training and are removed at inference time.

\section{Experiments}
\label{sec:experiments}
We evaluate \method on long-horizon interactive decision making with two complementary goals. First, we ask whether learning an executable oracle continuation improves the policy beyond outcome-based RL and single-step oracle supervision. Second, we isolate whether the gain comes from the executable oracle continuation or from policy-reachability weighting.

\subsection{Experimental Setup}
\paragraph{Benchmarks and baselines.}
ALFWorld evaluates multi-step household instruction following in an interactive text environment~\citep{shridhar2021alfworld}. We report success on the in-distribution and out-of-distribution splits (\emph{In-Success} and \emph{Out-Success}). WebShop requires an agent to search, navigate product pages, select attributes, and purchase an item~\citep{yao2022webshop}; following prior work, we report both task score and strict task success rate. We compare against prompting baselines (Qwen2.5, ReAct~\citep{yao2023react}, and Reflexion~\citep{shinn2023reflexion}), critic-based PPO~\citep{schulman2017ppo}, group-based RLOO~\citep{ahmadian2024rloo} and GRPO~\citep{shao2024deepseekmath}, and recent long-horizon agentic RL methods GiGPO~\citep{feng2025gigpo}, GraphGPO~\citep{cheng2026graphgpo} and HGPO~\citep{he2026hgpo}.

\paragraph{Models and training.}
Our main experiments use Qwen2.5-1.5B-Instruct and Qwen2.5-7B-Instruct. We use the ALFWorld expert planner and a goal-metadata oracle for WebShop as training-time oracles. An eligible group contains at least one failed rollout, from which one pre-action state is sampled as the oracle root. Unless otherwise noted, \method uses one root per eligible group and $\lambda_O=0.1$. Detailed settings are given in Appendix~\ref{app:config}.

\subsection{Main Results}
Table~\ref{tab:main} summarizes results on ALFWorld and WebShop. Across both model scales, \method improves ALFWorld and WebShop success on both RL backbones, with the largest gains on out-of-distribution metrics. At 1.5B, GraphGPO+\method improves ALFWorld seen/unseen success from $90.63/88.02$ to $95.83/91.67$ and WebShop score/success from $87.27/77.93$ to $88.90/79.27$; HGPO+\method improves ALFWorld from $92.19/87.24$ to $94.53/90.10$ and WebShop success from $73.70$ to $77.47$. At 7B, GraphGPO+\method improves ALFWorld from $91.67/76.30$ to $95.57/87.50$ and WebShop from $89.77/80.47$ to $91.93/83.67$, while HGPO+\method improves ALFWorld from $94.01/90.89$ to $94.53/91.41$ and WebShop success from $74.47$ to $79.63$. The largest gain is $11.20$ points on ALFWorld unseen with 7B GraphGPO, showing a particularly large benefit on out-of-distribution evaluation for this backbone.


\begin{table}[t]
\caption{Main results on ALFWorld and WebShop. ALFWorld reports in-distribution and out-of-distribution success rates (\%); WebShop reports task score and task success rate (\%). Results are averaged over 3 random seeds during testing. The best and second-best performances are highlighted in bold and underlined, respectively.}
\label{tab:main}
\centering
\small
\setlength{\tabcolsep}{4.0pt}
\resizebox{\textwidth}{!}{%
\begin{tabular}{lllcccc}
\toprule
Model & Type & Method & \multicolumn{2}{c}{ALFWorld} & \multicolumn{2}{c}{WebShop}\\
\cmidrule(lr){4-5}\cmidrule(lr){6-7}
& & & In-Success & Out-Success & Task Score & Task Success\\
\midrule
\multirow{2}{*}{Closed-source}
& Prompting & GPT-4o & 48.0 & 46.0 & 31.8 & 23.7\\
& Prompting & Gemini-2.5-Pro & 60.3 & 50.5 & 42.5 & 35.9\\
\midrule
\multirow{11}{*}{Qwen2.5-1.5B-Instruct}
& Prompting & Qwen2.5 & 4.1 & -- & 23.1 & 5.2\\
& Prompting & ReAct & 12.8 & -- & 40.1 & 11.3\\
& Prompting & Reflexion & 21.8 & -- & 55.8 & 21.9\\
& RL Training & PPO (with critic) & \result{54.4}{3.1} & -- & \result{73.8}{3.0} & \result{51.5}{2.9}\\
& RL Training & RLOO & \result{69.7}{2.5} & \result{68.7}{10.7} & \result{73.9}{5.6} & \result{52.1}{6.7}\\
& RL Training & GRPO & \result{72.8}{3.6} & \result{70.1}{2.5} & \result{75.8}{3.5} & \result{56.8}{3.8}\\
& RL Training & GiGPO & \result{88.27}{1.08} & \result{86.47}{2.25} & \result{83.10}{1.6} & \result{65.00}{3.21}\\
\cmidrule(lr){2-7}
& RL Training & GraphGPO & \result{90.63}{3.40} & \result{88.02}{3.23} & \result{87.27}{2.83} & \secondresult{77.93}{2.90}\\
\rowcolor{gray!10}
& RL Training & GraphGPO + \method & \bestresult{95.83}{1.22} & \bestresult{91.67}{0.92} & \bestresult{88.90}{1.59} & \bestresult{79.27}{1.72}\\
& RL Training & HGPO & \result{92.19}{0.80} & \result{87.24}{4.01} & \result{87.80}{1.67} & \result{73.70}{2.95}\\
\rowcolor{gray!10}
& RL Training & HGPO + \method & \secondresult{94.53}{1.33} & \secondresult{90.10}{1.61} & \secondresult{88.43}{0.93} & \result{77.47}{1.88}\\
\midrule
\multirow{11}{*}{Qwen2.5-7B-Instruct}
& Prompting & Qwen2.5 & 14.8 & -- & 26.4 & 7.8\\
& Prompting & ReAct & 31.2 & -- & 46.2 & 19.5\\
& Prompting & Reflexion & 42.7 & -- & 58.1 & 28.8\\
& RL Training & PPO (with critic) & \result{77.08}{1.12} & \result{76.23}{1.46} & \result{81.4}{3.1} & \result{68.7}{5.1}\\
& RL Training & RLOO & \result{77.86}{0.03} & \result{73.95}{0.05} & \result{80.3}{3.2} & \result{65.7}{4.0}\\
& RL Training & GRPO & \result{78.64}{0.73} & \result{76.82}{1.47} & \result{79.3}{2.8} & \result{66.1}{3.7}\\
& RL Training & GiGPO & \result{92.29}{1.56} & \result{86.94}{2.08} & \result{79.31}{2.80} & \result{66.10}{3.71}\\
\cmidrule(lr){2-7}
& RL Training & GraphGPO & \result{91.67}{0.92} & \result{76.30}{2.76} & \secondresult{89.77}{0.81} & \secondresult{80.47}{1.53}\\
\rowcolor{gray!10}
& RL Training & GraphGPO + \method & \bestresult{95.57}{0.46} & \result{87.50}{2.06} & \bestresult{91.93}{0.42} & \bestresult{83.67}{0.95}\\
& RL Training & HGPO & \result{94.01}{2.05} & \secondresult{90.89}{1.97} & \result{87.83}{2.00} & \result{74.47}{3.10}\\
\rowcolor{gray!10}
& RL Training & HGPO + \method & \secondresult{94.53}{3.15} & \bestresult{91.41}{2.12} & \result{87.63}{1.16} & \result{79.63}{0.50}\\
\bottomrule
\end{tabular}%
}
\end{table}

\subsection{Ablation of Oracle Supervision}

We ablate two components of \method: whether supervision extends beyond the root oracle action, and whether downstream oracle actions are weighted by prefix reachability. All variants use Qwen2.5-1.5B-Instruct with GraphGPO.

\emph{Root-only} uses only the first oracle action and discards the remaining continuation. \emph{Full-Flat} instead trains the complete oracle continuation, but assigns every step in a branch the same average weight,
\begin{equation}
 \bar w_r=\frac{1}{L_r}\sum_{k=0}^{L_r-1}w_{r,k},
 \qquad
 \mathcal L_{\mathrm{flat},r}^{\mathrm{branch}}
 =g_r\bar w_r\sum_{k=0}^{L_r-1}\ell_{r,k}.
\label{eq:flat}
\end{equation}
This construction preserves the total branch-level weight, $L_r\bar w_r=\sum_{k=0}^{L_r-1}w_{r,k}$, while removing its step-wise allocation across oracle depth. Full-Flat retains a policy-dependent branch scale through $\bar w_r$; this control tests how that weight is distributed across oracle steps. The full \method trains the same multi-step oracle continuation while retaining the exclusive-prefix reachability weight $w_{r,k}$ for each downstream action.

\begin{table}[t]
\caption{Ablation of oracle supervision design on ALFWorld and WebShop. Results use Qwen2.5-1.5B-Instruct with GraphGPO.
Root-only supervises only the first oracle action, Full-Flat uses the full oracle continuation with a shared branch-level weight, and \method applies exclusive-prefix reachability weighting along the continuation.}
\label{tab:ablation}
\centering
\small
\setlength{\tabcolsep}{5.0pt}
\begin{tabular}{lllcccc}
\toprule
& & & \multicolumn{2}{c}{ALFWorld} & \multicolumn{2}{c}{WebShop}\\
\cmidrule(lr){4-5}\cmidrule(lr){6-7}
Variant & Oracle target & Weighting & Seen SR & Unseen SR & Task Score & Task Success\\
\midrule
Base$^{\dagger}$ & None & --- & \result{90.63}{3.40} & \result{88.02}{3.23} & \result{87.27}{2.83} & \result{77.93}{2.90}\\
+ Root-only & Root action & --- & \result{87.76}{1.67} & \result{83.85}{2.76} & \result{87.30}{1.54} & \result{77.73}{2.38}\\
+ Full-Flat & Full continuation & Branch mean & \result{94.79}{0.46} & \result{88.28}{2.35} & \result{87.63}{1.91} & \result{78.33}{2.75}\\
+ \method & Full continuation & Exclusive prefix & \result{95.83}{1.22} & \result{91.67}{0.92} & \result{88.90}{1.59} & \result{79.27}{1.72}\\
\bottomrule
\end{tabular}
\end{table}

\paragraph{Effect of the oracle continuation.}
Root-only does not improve over GraphGPO and instead decreases ALFWorld seen/unseen success by 2.87/4.17 points. It also yields little change on WebShop. In contrast, Full-Flat substantially improves the results, reaching $94.79$ seen success on ALFWorld and $78.33$ task success on WebShop. Relative to Root-only, using the full executable continuation improves ALFWorld seen/unseen success by $7.03/4.43$ points and WebShop task score/success by $0.33/0.60$ points. These results indicate that useful oracle supervision extends beyond the first corrective action.

\paragraph{Effect of prefix reachability weighting.}

Replacing the shared branch weight with the exclusive-prefix weights further improves all four metrics. Compared with Full-Flat, \method gains $1.04$ points on ALFWorld seen success and $3.39$ points on unseen success, while WebShop task score and task success improve by $1.27$ and $0.94$ points. The larger gain on ALFWorld unseen suggests that selectively down-weighting deep oracle actions whose preceding prefix is poorly supported by the current policy may be particularly beneficial for out-of-distribution generalization. To isolate the weighting rule from differences in collected trajectories, we further replay all weighting variants on the same frozen oracle bank (Appendix~\ref{app:fixed_bank}). Under identical roots and oracle continuations, \method achieves $88.02/85.94$ seen/unseen success, compared with $85.42/84.64$ for flat within-branch weighting and $87.24/82.55$ for a depth-decay baseline, supporting the benefit of policy-dependent prefix reachability beyond depth decay alone.

\paragraph{Sensitivity to $\lambda_O$.}
\begin{wrapfigure}{r}{0.6\textwidth}
    \centering
    \vspace{-0.8em}
    \includegraphics[width=\linewidth]{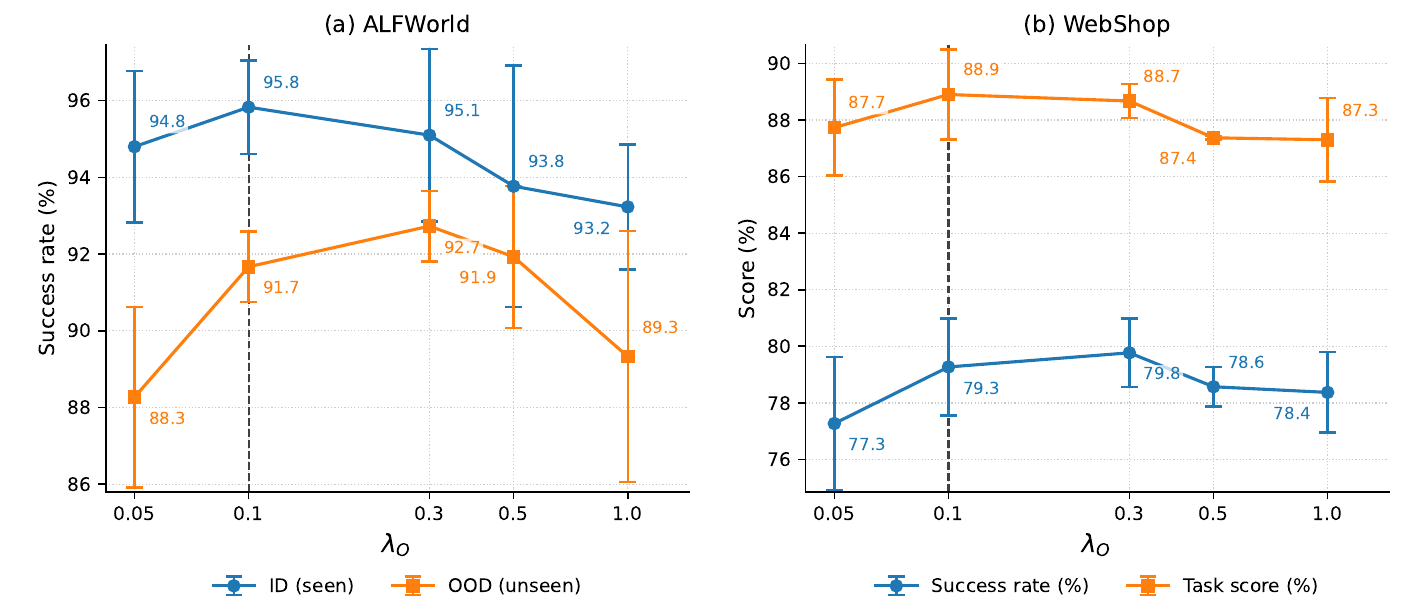}
    \caption{Sensitivity to $\lambda_O$ on ALFWorld and WebShop. Performance is stable over moderate oracle-loss weights, with $\lambda_O=0.1$ used as the default.}
    \label{fig:lambda_sensitivity}
    \vspace{-0.8em}
\end{wrapfigure}

Figure~\ref{fig:lambda_sensitivity} shows that \method is relatively stable over a moderate range of oracle-loss weights. On ALFWorld, both seen and unseen performance remain strong for $\lambda_O\in[0.1,0.5]$, while WebShop shows a similar pattern for task score and success rate. Performance drops when the oracle loss is too strong at $\lambda_O=1.0$, whereas $\lambda_O=0.05$ yields weaker results, particularly on unseen ALFWorld tasks. We therefore use $\lambda_O=0.1$ as the default setting in the main experiments.

\subsection{Policy Reachability Across Oracle Depth}

\begin{wrapfigure}{l}{0.65\textwidth}
    \centering
    \vspace{-0.8em}
    \includegraphics[width=\linewidth]{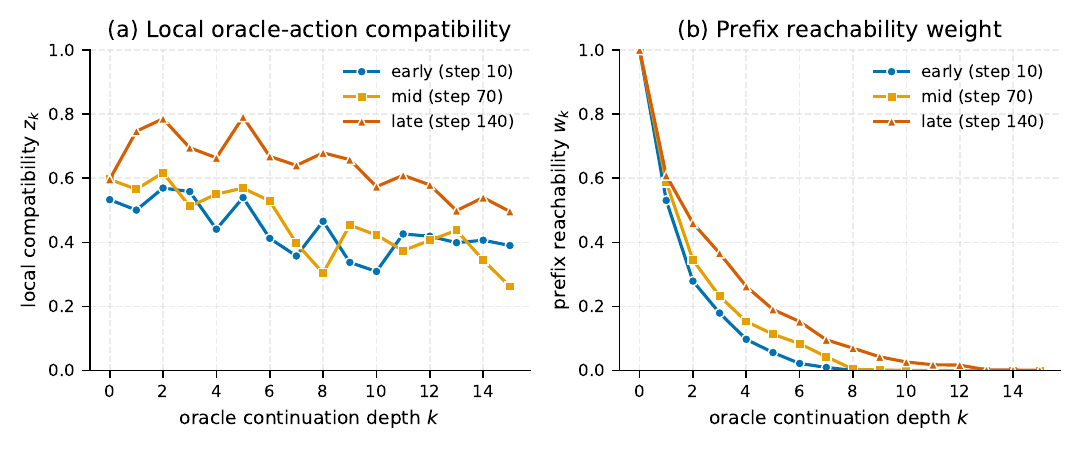}
    \caption{Policy compatibility along oracle continuations.
    Local compatibility $z_k$ and prefix reachability weight $w_k$ for the Qwen2.5-1.5B GraphGPO+$\method$ model on ALFWorld at training steps 10, 70, and 140.}
    \label{fig:depth}
    \vspace{-0.8em}
\end{wrapfigure}

Figure~\ref{fig:depth} shows why local compatibility alone is insufficient for weighting a long oracle continuation. In Figure~\ref{fig:depth}(a), $z_k$ fluctuates with depth and often remains relatively high even for later oracle actions. This is because $z_k$ only measures whether the action at the current oracle state is compatible with the behavior policy; it does not account for whether the policy could follow the preceding oracle actions to reach that state. Figure~\ref{fig:depth}(b) captures this missing information through the prefix weight $w_k=\prod_{j<k}z_j$. Once an earlier oracle action has low compatibility, the weights of all subsequent actions are reduced accordingly. As a result, $w_k$ drops quickly at the early checkpoint, while the later checkpoint maintains larger weights over a longer portion of the continuation. As training progresses, \method therefore makes use of increasingly deeper oracle supervision, while avoiding assigning the same weight to every downstream oracle action.

\section{Conclusion}

We presented \method, a framework that incorporates executable oracle continuations into multi-turn agentic RL. Starting from states visited by the current policy, \method restores the environment, executes oracle continuations, and uses the resulting action sequences as auxiliary supervision. To better utilize long-horizon guidance, we introduce policy-reachability weighting, which adjusts oracle supervision based on the compatibility between oracle prefixes and the current policy. With Qwen2.5-1.5B and 7B models, adding \method improves ALFWorld success and WebShop strict success for both GraphGPO and HGPO. Ablations show that learning from full executable continuations provides substantially stronger gains than isolated corrective actions, while reachability weighting further improves generalization. These results highlight the value of executable multi-step oracle experience and the importance of accounting for policy reachability when extending supervision beyond immediate actions.


\bibliography{references}
\bibliographystyle{iclr2027_conference}

\clearpage
\appendix
\section{Additional Controlled Results}
\label{app:additional_results}

\subsection{Fixed-Bank Weighting Control}
\label{app:fixed_bank}

The online ablation in Table~\ref{tab:ablation} compares variants that continue to collect their own training trajectories. We therefore construct a frozen oracle bank from the Qwen2.5-1.5B GraphGPO checkpoint at step 50 and replay all weighting variants on exactly the same supervision data. The bank contains 271 valid roots and 5,002 oracle-continuation steps, and all variants use the same roots, continuations, optimizer settings, and step-50-to-80 replay schedule. The stored reasoning contexts, token masks, compatibility scores, prefix weights, and branch gates remain fixed throughout replay. All variants share the auxiliary sampling schedule while continuing ordinary on-policy GraphGPO updates.

The \emph{Flat} variant distributes each branch's total supervision mass uniformly across its oracle steps. \emph{Depth decay} replaces policy reachability with a fixed geometric decay, using $0<\alpha_{\rm depth}\leq1$,
\begin{equation}
\widetilde w_{r,k}=\kappa_r\alpha_{\rm depth}^{k},
\qquad
\kappa_r=\frac{\sum_{j=0}^{L_r-1} w_{r,j}}
{\sum_{j=0}^{L_r-1}\alpha_{\rm depth}^{j}},
\label{eq:depth_only}
\end{equation}
while preserving the same total branch weight as \method. Before replay, we fit $\log\alpha_{\rm depth}$ once by least squares through the origin to $\log w_{r,k}$ versus $k$, using valid positive-gate branches with $k\geq1$ and $w_{r,k}>0$; $\alpha_{\rm depth}$ is clipped to $[10^{-4},1]$ and then held fixed. \emph{Shuffled} keeps the same set of SERO weights within each branch but permutes their positions using a fixed, reproducible permutation for that branch, breaking the correspondence between prefix compatibility and the supervised oracle step.

Table~\ref{tab:fixed_bank} shows that \method performs best under this controlled replay setting, reaching $88.02/85.94$ seen/unseen success. It improves over Flat by $2.60/1.30$ points and over Depth decay by $0.78/3.39$ points. More importantly, shuffling the SERO weights reduces performance to $85.16/80.73$, a drop of $2.86$ points on seen tasks and $5.21$ points on unseen tasks. This shows that the benefit does not come only from the overall amount of supervision or from assigning smaller weights to deeper steps; the weight needs to remain aligned with the reachability of the corresponding oracle prefix.

\subsection{Matched Extra-Exploration Control}
\label{app:matched_budget}

We further test whether the gain of \method can be explained by its additional environment interactions. The baseline trains GraphGPO for 15 extra steps without oracle supervision, targeting the estimated environment-transition budget of GraphGPO+$\method$. The estimate includes primary rollouts, prefix replay for restoration, and oracle continuations, using calibration windows from early, middle, and late training; the match is approximate. As shown in Table~\ref{tab:matched_budget}, the extra exploration improves GraphGPO from $90.63/88.02$ to $93.49/90.36$ seen/unseen success. However, \method still reaches $95.83/91.67$, outperforming the matched baseline by $2.34$ and $1.31$ points, respectively. The extended GraphGPO run narrows the gap but does not match \method in this control.

\par
\medskip
\noindent
\begin{minipage}[t]{0.47\textwidth}
\vspace{0pt}
\centering
\small
\captionof{table}{Fixed-bank weighting control on ALFWorld.
All variants replay the same frozen roots and oracle continuations from the GraphGPO step-50 checkpoint and are evaluated at step 80. Shuffled preserves the SERO weights within each branch but permutes their positions along the continuation.}
\label{tab:fixed_bank}
\vspace{0.4em}
\setlength{\tabcolsep}{4.5pt}
\begin{tabular}{lcc}
\toprule
Weighting & Seen SR & Unseen SR\\
\midrule
Flat & \result{85.42}{2.39} & \result{84.64}{1.79}\\
Depth decay & \result{87.24}{2.94} & \result{82.55}{0.46}\\
\method & \bestresult{88.02}{1.22} & \bestresult{85.94}{0.75}\\
Shuffled & \result{85.16}{3.43} & \result{80.73}{1.79}\\
\bottomrule
\end{tabular}
\end{minipage}
\hfill
\begin{minipage}[t]{0.47\textwidth}
\vspace{0pt}
\centering
\small
\captionof{table}{Matched-interaction control on ALFWorld.
Extra denotes GraphGPO trained for 15 additional steps to approximately match the estimated environment-transition budget of GraphGPO+$\method$, including prefix restoration and oracle execution.}
\label{tab:matched_budget}
\vspace{0.4em}
\setlength{\tabcolsep}{3.5pt}
\begin{tabular}{lcc}
\toprule
Method & Seen SR & Unseen SR\\
\midrule
GraphGPO
& \result{90.63}{3.40}
& \result{88.02}{3.23}\\
GraphGPO + Extra
& \result{93.49}{1.22}
& \result{90.36}{3.43}\\
GraphGPO + \method
& \result{95.83}{1.22}
& \result{91.67}{0.92}\\
\bottomrule
\end{tabular}
\end{minipage}
\par
\medskip

\subsection{Empirical Check of the Reachability Proxy}
\label{app:reachability_check}

We test whether $w_k$ tracks the frozen policy's ability to reproduce an oracle prefix. Using the GraphGPO step-50 checkpoint and the 271-root bank of Appendix~\ref{app:fixed_bank}, we sample 16 commands at each recorded oracle state and run eight independent prefix trials per root. A trial proceeds along the recorded branch only while its sampled commands exactly match the oracle commands; it does not execute a new environment trajectory. The single-step compatibility $z_k$ correlates with the sampled action-match frequency (Spearman $\rho=0.763$). Table~\ref{tab:reachability_check} groups the observed prefix-match frequency by $w_k$ across root--depth pairs. The observed match frequency rises across the $w_k$ bins. Since $z_k$ is a geometric mean over command-token probabilities, $w_k$ ranks prefix compatibility but is not the probability of reproducing the full command prefix.

\begin{table}[H]
\centering
\small
\caption{Empirical oracle-prefix reproduction under the frozen GraphGPO step-50 policy. Bins pool 2,869 root--depth pairs from 271 roots; each root has eight sampled prefix trials. The final column gives the fraction of trials that reproduce the full oracle prefix.}
\label{tab:reachability_check}
\setlength{\tabcolsep}{7pt}
\begin{tabular}{lrc}
\toprule
$w_k$ interval & Root--depth pairs & Prefix match (\%)\\
\midrule
$[0,0.05)$ & 2,449 & 0.00\\
$[0.05,0.10)$ & 75 & 1.00\\
$[0.10,0.25)$ & 127 & 1.08\\
$[0.25,0.50)$ & 92 & 6.66\\
$[0.50,1]$ & 126 & 40.97\\
\bottomrule
\end{tabular}
\end{table}

\subsection{Current-Policy Root Control}
\label{app:root_source}

We also compare the source of the restored oracle root under a fixed-bank replay from the GraphGPO step-50 checkpoint. The reference bank selects failed-rollout states visited by that checkpoint; the alternative selects states visited by a step-10 policy. Both matched banks contain 271 roots and 78 positive-gate branches, with 5,002 and 4,961 oracle steps, respectively; 242 of 271 paired roots have the same task family. After the same 30-update replay, the step-50 roots reach $88.0/85.9$ seen/unseen success, compared with $84.4/83.9$ for step-10 roots. The more recent roots perform better in this replay, although the task instances and root horizons are not exactly matched.

\subsection{ALFWorld Task-Type Breakdown}
\label{app:task_breakdown}
Table~\ref{tab:alfworld_task_breakdown} breaks ALFWorld success into the six task families used by the benchmark. We report GraphGPO and HGPO at both 1.5B and 7B under the same 128-game final-checkpoint protocol as the main table. Improvements with \method appear across several task families on both backbones and model scales, showing that the aggregate gains are not driven by a single category.

\begin{table}[H]
\centering
\small
\caption{ALFWorld success rate by task type.
Pick2 denotes tasks requiring two objects to be placed. All reports the overall success rate on the corresponding split.}
\label{tab:alfworld_task_breakdown}
\setlength{\tabcolsep}{3.2pt}
\resizebox{\textwidth}{!}{%
\begin{tabular}{lllcccccc|c}
\toprule
Model & Split & Method & Pick & Clean & Cool & Look & Heat & Pick2 & All\\
\midrule
\multirow{8}{*}{1.5B}
& Seen & GraphGPO
& \result{92.57}{3.65} & \result{100.00}{0.00} & \result{81.27}{4.41} & \result{81.73}{8.55} & \result{87.17}{8.04} & \result{87.27}{8.49} & \result{90.63}{3.40}\\
& Seen & GraphGPO + \method
& \result{99.17}{1.44} & \result{96.00}{1.13} & \result{87.33}{4.18} & \result{95.23}{8.26} & \result{98.13}{3.23} & \result{96.47}{3.07} & \result{95.83}{1.22}\\
& Seen & HGPO
& \result{93.27}{1.00} & \result{97.10}{2.69} & \result{83.77}{3.03} & \result{77.13}{4.97} & \result{98.33}{2.89} & \result{92.77}{8.57} & \result{92.19}{0.80}\\
& Seen & HGPO + \method
& \result{100.00}{0.00} & \result{97.10}{2.69} & \result{85.33}{1.88} & \result{91.90}{7.34} & \result{96.50}{6.06} & \result{89.23}{5.59} & \result{94.53}{1.33}\\
\cmidrule(lr){2-10}
& Unseen & GraphGPO
& \result{79.77}{9.08} & \result{85.33}{5.47} & \result{96.03}{3.86} & \result{100.00}{0.00} & \result{85.03}{6.66} & \result{87.70}{14.71} & \result{88.02}{3.23}\\
& Unseen & GraphGPO + \method
& \result{83.73}{1.70} & \result{88.60}{7.85} & \result{100.00}{0.00} & \result{100.00}{0.00} & \result{94.60}{0.35} & \result{85.87}{5.34} & \result{91.67}{0.92}\\
& Unseen & HGPO
& \result{79.63}{6.30} & \result{84.40}{9.26} & \result{84.87}{4.51} & \result{100.00}{0.00} & \result{94.60}{0.35} & \result{94.07}{5.59} & \result{87.24}{4.01}\\
& Unseen & HGPO + \method
& \result{87.07}{9.35} & \result{93.83}{4.27} & \result{92.43}{2.24} & \result{96.83}{5.48} & \result{85.90}{7.54} & \result{88.07}{5.38} & \result{90.10}{1.61}\\
\midrule
\multirow{8}{*}{7B}
& Seen & GraphGPO
& \result{99.17}{1.44} & \result{97.10}{2.69} & \result{77.57}{3.56} & \result{100.00}{0.00} & \result{86.20}{7.41} & \result{87.57}{5.98} & \result{91.67}{0.92}\\
& Seen & GraphGPO + \method
& \result{100.00}{0.00} & \result{100.00}{0.00} & \result{85.73}{3.15} & \result{83.80}{10.74} & \result{93.17}{5.92} & \result{100.00}{0.00} & \result{95.57}{0.46}\\
& Seen & HGPO
& \result{99.17}{1.44} & \result{100.00}{0.00} & \result{85.90}{6.63} & \result{84.67}{5.77} & \result{87.70}{3.08} & \result{98.13}{3.23} & \result{94.01}{2.05}\\
& Seen & HGPO + \method
& \result{100.00}{0.00} & \result{95.37}{5.36} & \result{91.03}{8.71} & \result{84.90}{7.53} & \result{96.57}{2.98} & \result{91.10}{2.91} & \result{94.53}{3.15}\\
\cmidrule(lr){2-10}
& Unseen & GraphGPO
& \result{72.87}{8.41} & \result{76.87}{8.95} & \result{83.60}{5.16} & \result{96.53}{3.03} & \result{69.47}{4.79} & \result{68.00}{8.63} & \result{76.30}{2.76}\\
& Unseen & GraphGPO + \method
& \result{68.50}{4.35} & \result{83.03}{7.20} & \result{100.00}{0.00} & \result{100.00}{0.00} & \result{92.77}{3.36} & \result{89.97}{9.65} & \result{87.50}{2.06}\\
& Unseen & HGPO
& \result{84.43}{8.17} & \result{93.83}{4.27} & \result{92.43}{2.24} & \result{98.13}{3.23} & \result{80.37}{8.36} & \result{100.00}{0.00} & \result{90.89}{1.97}\\
& Unseen & HGPO + \method
& \result{85.23}{5.50} & \result{93.83}{4.27} & \result{92.43}{2.24} & \result{100.00}{0.00} & \result{90.90}{6.59} & \result{94.43}{5.10} & \result{91.41}{2.12}\\
\bottomrule
\end{tabular}%
}
\end{table}

For GraphGPO, the largest 1.5B seen gains appear on Look, Heat, and Pick2, while unseen gains are strongest on Heat; at 7B, the largest unseen gains occur on Clean, Cool, Heat, and Pick2. Its few category-level regressions are not consistent across model scales. HGPO likewise shows gains across multiple categories, with some task-specific trade-offs.

\subsection{Training Dynamics}
\label{app:training_dynamics}

\begin{figure}[t]
    \centering
    \includegraphics[width=0.92\linewidth]{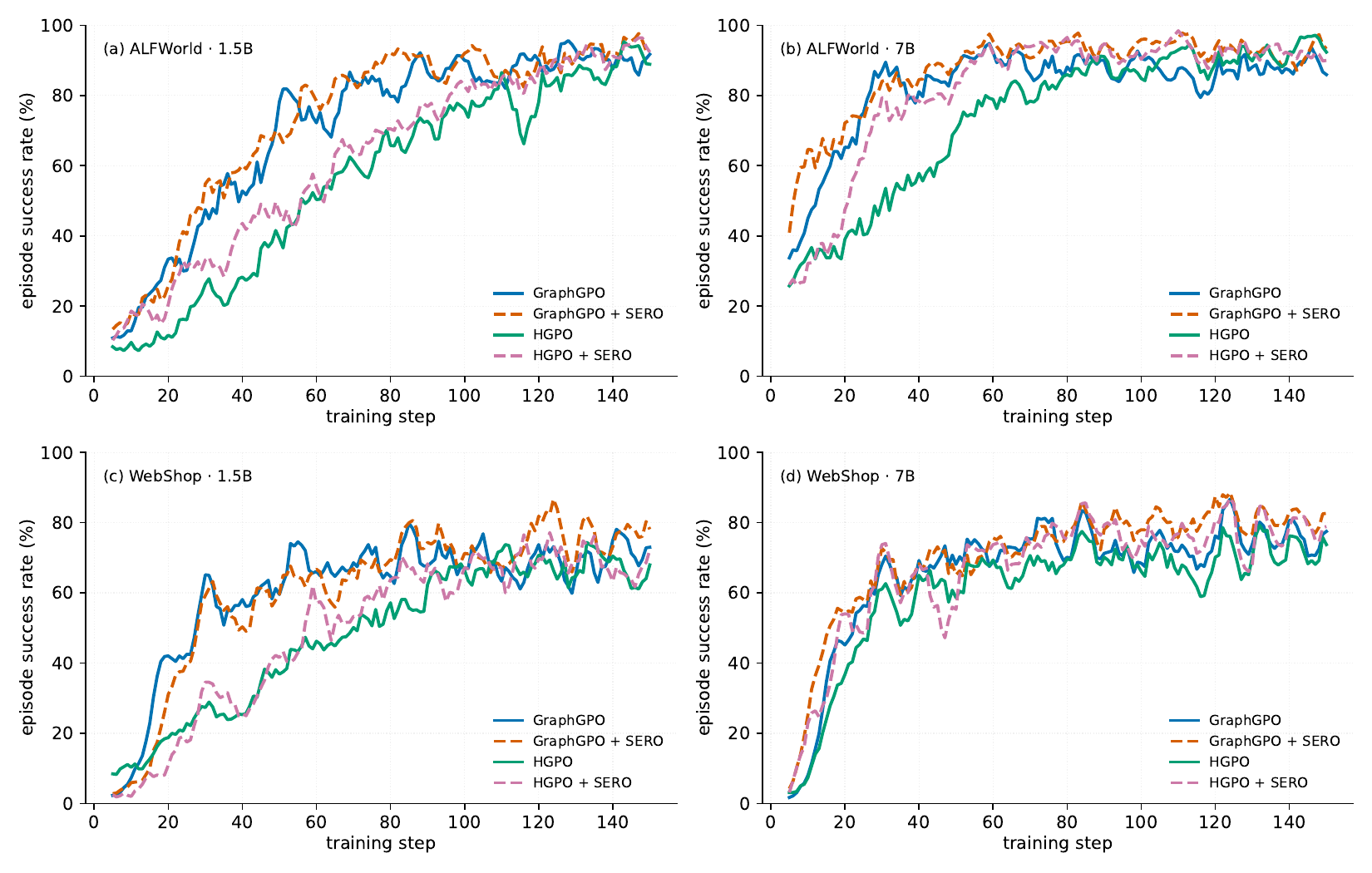}
    \caption{Training success rate on ALFWorld and WebShop.
    We compare GraphGPO and HGPO with and without \method for Qwen2.5-1.5B-Instruct and Qwen2.5-7B-Instruct. Each curve reports the episode success rate of the primary training rollouts over training steps.}
    \label{fig:training_curves}
\end{figure}

Figure~\ref{fig:training_curves} compares the training success rates of GraphGPO and HGPO with and without \method across both model scales and benchmarks. On ALFWorld, adding \method generally leads to faster improvement during the early and middle stages of training, with the difference particularly clear for HGPO at 1.5B and for both backbones at 7B. The WebShop curves show more crossings: for example, the 1.5B GraphGPO baseline improves earlier during part of the initial training period, while the advantage of GraphGPO+\method is clearer later. The curves therefore show gains before the final evaluation checkpoint, with their timing and magnitude varying across settings.

\subsection{Short-Bridge Control}
\label{app:short_bridge}

The short-bridge control executes one oracle action from the restored root and then returns control to the frozen behavior policy for the remaining horizon. Its gate is computed from the resulting bridge return, $g_r^{\rm bridge}=v_r[G_r^{\rm bridge}-G_r^\pi]_+$, but only the root oracle action is distilled:
\begin{equation}
    \mathcal L_{\rm bridge}
    =\frac{1}{\max(N_{\rm root},1)}
      \sum_{r=1}^{N_{\rm root}}\operatorname{sg}[g_r^{\rm bridge}]\,
      \ell_{r,0}(\theta).
\label{eq:short_bridge}
\end{equation}
The behavior-policy continuation provides no auxiliary target and does not enter the base RL statistics. This isolates whether a single corrective action followed by the current policy is sufficient, or whether learning the full oracle continuation provides additional value.

\begin{table}[H]
\centering
\small
\caption{\textbf{Short-bridge control on ALFWorld and WebShop.}
Root-only supervises only the first oracle target, Short Bridge supervises that root action while using a frozen-policy continuation only to evaluate its outcome, and \method learns the full executable oracle continuation.}
\label{tab:short_bridge}
\setlength{\tabcolsep}{5.0pt}
\begin{tabular}{lcccc}
\toprule
& \multicolumn{2}{c}{ALFWorld} & \multicolumn{2}{c}{WebShop}\\
\cmidrule(lr){2-3}\cmidrule(lr){4-5}
Variant & Seen SR & Unseen SR & Task Score & Task Success\\
\midrule
GraphGPO
& \result{90.63}{3.40}
& \result{88.02}{3.23}
& \result{87.27}{2.83}
& \result{77.93}{2.90}\\
+ Root-only
& \result{87.76}{1.67}
& \result{83.85}{2.76}
& \result{87.30}{1.54}
& \result{77.73}{2.38}\\
+ Short Bridge
& \result{91.67}{5.83}
& \result{86.72}{2.35}
& \result{87.60}{1.47}
& \result{78.77}{1.75}\\
+ \method
& \result{95.83}{1.22}
& \result{91.67}{0.92}
& \result{88.90}{1.59}
& \result{79.27}{1.72}\\
\bottomrule
\end{tabular}
\end{table}

As shown in Table~\ref{tab:short_bridge}, Short Bridge consistently improves over Root-only on both benchmarks. On ALFWorld, it raises seen/unseen success from $87.76/83.85$ to $91.67/86.72$, but still trails the full \method by $4.16/4.95$ points. On WebShop, Short Bridge improves task score/success from $87.30/77.73$ to $87.60/78.77$, while full \method further reaches $88.90/79.27$. These results show that a one-step oracle intervention is helpful, but learning the full executable continuation provides a stronger training signal.

\subsection{Horizon-Stratified Evaluation}
\label{app:horizon}

We further group ALFWorld episodes by expert-plan length to examine how the effect of \method changes with task horizon. Table~\ref{tab:horizon} reports results for short ($\leq5$), medium ($6$), and long ($\geq7$) plans using the same final checkpoints as the main evaluation.

\begin{table}[H]
\centering
\small
\caption{\textbf{ALFWorld success stratified by expert-plan length.}
Short, medium, and long correspond to expert-plan lengths $\leq5$, $6$, and $\geq7$, respectively.}
\label{tab:horizon}
\begin{tabular}{llccc}
\toprule
Method & Split & Short ($\leq5$) & Medium ($6$) & Long ($\geq7$)\\
\midrule
GraphGPO
& Seen
& \result{90.37}{1.59}
& \result{91.65}{4.34}
& \result{90.02}{7.14}\\
GraphGPO + \method
& Seen
& \result{98.65}{1.18}
& \result{98.37}{1.42}
& \result{90.14}{5.13}\\
GraphGPO
& Unseen
& \result{88.95}{3.14}
& \result{91.89}{3.63}
& \result{83.50}{6.16}\\
GraphGPO + \method
& Unseen
& \result{95.18}{2.17}
& \result{91.60}{1.51}
& \result{87.32}{1.39}\\
\bottomrule
\end{tabular}
\end{table}

On the seen split, \method improves short- and medium-horizon success by $8.28$ and $6.72$ points, while the long-horizon bucket remains nearly unchanged. On the unseen split, success improves by $6.23$ points for short plans and $3.82$ points for long plans, with similar performance on the medium bucket. These results show that the gains are not limited to short episodes: \method also improves the more challenging long-horizon unseen tasks, where GraphGPO increases from $83.50$ to $87.32$.

\section{Notation and Action Supervision}
\label{app:commands}
We use $s_t$ for the information available to the policy at decision step $t$: the current observation together with the interaction context exposed by the agent interface. Exact restoration may additionally use simulator state kept by the environment wrapper. Table~\ref{tab:notation} collects the notation used below.

\begin{table}[h]
\centering\small
\caption{Notation used in the method and experiments.}
\label{tab:notation}
\begin{tabular}{lp{0.76\linewidth}}
\toprule
Symbol & Meaning\\
\midrule
$s_t$ & Policy-visible state at step $t$, including the current observation and interaction context.\\
$s_k^O$ & State on the oracle branch at step $k$; $s_0^O$ is the recorded on-policy root.\\
$\pi_b,\pi_\theta$ & Frozen behavior policy and trainable policy.\\
$a_k^\star$ & Executed oracle action at oracle step $k$.\\
$\ell_k,z_k$ & Trainable action loss and detached behavior-policy compatibility.\\
$w_k$ & Policy-reachability proxy, $\prod_{j<k}z_j$.\\
$G_r^\pi,G_r^O$ & Policy and oracle returns from the same restored root and remaining horizon.\\
$g_r,v_r$ & Detached branch gate and branch-validity indicator.\\
$L_r,N_{\rm root}$ & Continuation length and number of attempted roots.\\
$c_R$ & Reward scale used to normalize the branch-return gap.\\
$\lambda_O$ & Weight of the auxiliary objective relative to the base RL loss.\\
$\operatorname{sg}[\cdot]$ & Stop-gradient operator.\\
\bottomrule
\end{tabular}
\end{table}

\subsection{Language Commands}
At oracle step $k$, let the planner command contain tokens $y_{k,1}^\star,\ldots,y_{k,m_k}^\star$. If the actor produces a reasoning prefix before the command, we denote the prefix sampled from the frozen behavior policy by $c_k$. Both behavior-policy scoring and the auxiliary loss use the same replay context $(s_k^O,c_k)$; for an action-only policy, $c_k$ is empty. We compute
\begin{align}
 \ell_k(\theta)
 &=-\frac{1}{m_k}\sum_{j=1}^{m_k}
 \log\pi_\theta(y_{k,j}^\star\mid s_k^O,c_k,y_{k,<j}^\star),
 \label{eq:commandnll}\\
 z_k
 &=\exp\!\left(\frac{1}{m_k}\sum_{j=1}^{m_k}
 \log\pi_b(y_{k,j}^\star\mid s_k^O,c_k,y_{k,<j}^\star)\right).
 \label{eq:commandcompat}
\end{align}
The same serialization and command-token mask are used for both terms. The sampled $c_k$ is detached replay context, and no marginalization over reasoning prefixes is attempted. The geometric mean in Equation~\ref{eq:commandcompat} therefore defines a length-normalized compatibility score rather than a calibrated marginal action probability or the probability of emitting the entire command string.

\subsection{Policy-Reachability Interpretation}
\label{app:interpretation}
For categorical actions in a deterministic environment, $\prod_{j<k}\pi_b(a_j^\star\mid s_j^O)$ equals the probability of reproducing that oracle action prefix. More generally, it is an action-compatibility score along the realized oracle state sequence. For tokenized commands, $w_k=\prod_{j<k}z_j$ is the analogous length-normalized compatibility proxy, not the probability of emitting the full command prefix. The exclusive prefix is important: $w_0=1$, and low compatibility at step $k$ reduces supervision only for later oracle actions through $w_{k+1}=w_kz_k$.

\section{Training Algorithm and Restoration}
\label{app:algorithm}
\begin{algorithm}[h]
\caption{One training iteration of \method. Primary on-policy rollouts define the base RL update; selected roots then seed executable oracle continuations used by the auxiliary objective.}
\label{alg:sero}
\begin{algorithmic}[1]
\Require Policy $\pi_\theta$, base RL learner, task batch, oracle, coefficient $\lambda_O$, reward scale $c_R>0$
\State Freeze behavior snapshot $\pi_b\leftarrow\pi_\theta$
\State Collect primary on-policy rollouts $\mathcal D_\pi$ and exact pre-action root records
\State Compute base advantages and statistics using only $\mathcal D_\pi$
\State Select eligible roots $\mathcal R$; set $N_{\rm root}\leftarrow|\mathcal R|$ and $S\leftarrow0$
\For{each attempted root $r\in\mathcal R$}
    \State Restore the environment and policy-visible context in an isolated copy
    \If{restoration fails}
        \State Record the failed attempt and continue
    \EndIf
    \State Set $G_r^\pi$ to the realized return of the original on-policy suffix from this root
    \State Execute one oracle continuation under the original remaining horizon
    \State Record policy-visible states $s_{r,k}^O$, executed oracle actions $a_{r,k}^{\star}$, and branch outcome
    \State Compute detached branch gate $g_r$ using Equation~\ref{eq:gate}
    \State Set $w\leftarrow1$
    \For{$k=0,\ldots,L_r-1$}
        \State Sample $c_{r,k}$ once from $\pi_b$ at $s_{r,k}^O$; use an empty prefix for action-only policies
        \State Store $(s_{r,k}^O,c_{r,k},a_{r,k}^{\star})$, its serialization, and the command-token mask
        \State Compute detached $z_{r,k}$ from the stored context and mask using Equation~\ref{eq:commandcompat}
        \State Compute $\ell_{r,k}(\theta)$ from the same stored context and mask using Equation~\ref{eq:commandnll}
        \State $S\leftarrow S+\operatorname{sg}[g_r]\,\operatorname{sg}[w]\,\ell_{r,k}(\theta)$
        \State $w\leftarrow w z_{r,k}$
    \EndFor
\EndFor
\State $\mathcal L_{\mathrm{SERO}}\leftarrow S/\max(N_{\rm root},1)$
\State Update $\theta$ with $\mathcal L_{\mathrm{base}}(\mathcal D_\pi)+\lambda_O\mathcal L_{\mathrm{SERO}}$
\end{algorithmic}
\end{algorithm}

\subsection{Root Records and Restoration}
A root record contains the task instance, action prefix, current observation, remaining interaction budget, legal-action interface, agent-side history, and the realized return of the original trajectory from that root. Environments with serializable state are restored directly; otherwise the adapter resets the task and replays the recorded prefix. After restoration, the policy-visible prompt and admissible-action interface must match the original pre-action record before the oracle branch is executed; otherwise the branch is marked invalid.

The oracle continuation maintains its own environment history after the restored root. Later observations are produced by the oracle branch itself rather than copied from the original on-policy suffix.

\subsection{Data and Gradient Separation}
Primary RL data determine reward normalization, group statistics, graph construction, advantages, and policy ratios. Each backbone retains its native token-level policy-ratio computation; oracle examples are replayed only through the command-token loss $\ell_{r,k}(\theta)$, while $g_r$, $z_{r,k}$, and $w_{r,k}$ are detached. We normalize by attempted roots rather than total oracle steps so each root is one branching unit while longer valid continuations can contribute additional supervised decisions. The behavior snapshot is fixed while roots are collected and oracle actions are scored, and $N_{\rm root}$ is preserved across optimizer microbatches.

Implementation tests cover base-only parity at $\lambda_O=0$, isolation between primary and oracle environments, exact root reconstruction, legal-action and horizon preservation, exclusive-prefix ordering, command-only masking, attempted-root normalization, and zero oracle access during evaluation.

\section{Task Protocols and Oracle Backends}
\label{app:protocols}
\subsection{ALFWorld}
We use the text interface of ALFWorld~\citep{shridhar2021alfworld}. For an oracle branch, the adapter reconstructs the selected on-policy prefix and verifies the reconstructed prompt and admissible actions. It then obtains commands from the environment's expert-plan interface (\texttt{extra.expert\_plan}), checks each command against the current admissible actions, and executes it through the ordinary environment interface. For each task group containing at least one failure, one failed rollout is selected and one of its pre-action turns is sampled uniformly as the root; all-success groups are skipped.

ALFWorld uses binary task success for the oracle gate, so Equation~\ref{eq:gate} becomes
\begin{equation}
 g_r=v_r\,[\operatorname{success}(\tau_r^O)-
        \operatorname{success}(\tau_r^\pi)]_+,
 \qquad [x]_+=\max(x,0).
\label{eq:alfworld_gate}
\end{equation}
A valid oracle branch therefore receives a positive gate only when it improves the failed policy rollout from the same restored root. GraphGPO and HGPO retain their original ALFWorld wrappers; within each backbone, the base and SERO variants share the same task manifest, prompt construction, action interface, and evaluation protocol.

\subsection{WebShop}
WebShop~\citep{yao2022webshop} requires search, navigation, option selection, and purchase actions. Root selection follows the ALFWorld rule: from each group with at least one failure, we sample one failed rollout and then one of its pre-action turns uniformly; all-success groups are skipped, and eligibility uses strict success rather than the dense task score. The training-time oracle is a deterministic rule-based policy using the target product identifier and required options from goal metadata. When search is needed, it tries the sanitized retrieval query, full product name, and its first eight words in order; it navigates to the target, selects the required options, and purchases it through the same legal \texttt{search}/\texttt{click} interface as the policy. Restoration reproduces the page/session state, navigation history, selected options, task identity, and observation renderer. We report both task score and strict success, while the oracle gate uses strict task success.

\section{Experimental Configuration}
\label{app:config}
\subsection{ALFWorld Training Configuration}
Table~\ref{tab:config} lists the ALFWorld settings used in the matched GraphGPO/SERO and HGPO/SERO comparisons. Within each backbone, the base and +SERO runs use the same task manifests, action interface, and decoding settings.
\begin{table}[H]
\centering\small
\caption{ALFWorld training and evaluation configuration. Backbone-specific settings are shown explicitly; all base/+SERO comparisons are matched within backbone.}
\label{tab:config}
\begin{tabular}{lp{0.62\linewidth}}
\toprule
Item & Setting \\
\midrule
Models & Qwen2.5-1.5B-Instruct and Qwen2.5-7B-Instruct.\\
Rollout batch & 16 task groups $\times$ 8 rollouts per group.\\
GraphGPO prompt / response limit & 2,048 / 512 tokens.\\
HGPO prompt / response limit & 4,096 / 512 tokens.\\
Environment horizon & 50 interaction steps.\\
Training / evaluation temperature & $1.0$ / $0.4$.\\
Actor learning rate & $10^{-6}$.\\
Reference KL coefficient & $0.01$.\\
Invalid-action penalty & Coefficient $0.1$.\\
GraphGPO credit settings & Graph discount $0.10$; episode and step weights $1$; mean/std normalization.\\
HGPO credit settings & Discount $0.95$; length weighting with $\alpha_{\rm HGPO}=1.0$; mean/std normalization; base-group disabled.\\
Default SERO coefficient & $\lambda_O=0.1$.\\
Oracle roots & One root per eligible task group.\\
Compatibility score & Geometric mean of frozen behavior-policy probabilities over oracle command tokens.\\
Auxiliary target & Command-token NLL.\\
Final checkpoints & GraphGPO step 150; HGPO step 150.\\
Training hardware & 2 GPUs for 1.5B and 4 GPUs for 7B.\\
\bottomrule
\end{tabular}
\end{table}

\subsection{WebShop Training Configuration}
Table~\ref{tab:webshop_config} summarizes the backbone-specific WebShop settings. Within each backbone, the base and +SERO runs keep the same horizon, prompt format, action interface, and final-checkpoint evaluation protocol.
\begin{table}[H]
\centering\small
\caption{Backbone-specific WebShop settings used in the controlled comparisons.}
\label{tab:webshop_config}
\begin{tabular}{lcc}
\toprule
Setting & GraphGPO & HGPO \\
\midrule
Task groups $\times$ rollouts & $16\times8$ & $16\times8$ \\
Environment horizon & 15 & 30 \\
Prompt / response limit & 5,120 / 512 & 4,096 / 512 \\
PPO minibatch size & 64 & 64 \\
Backbone discount & $0.20$ & $0.95$ \\
Backbone normalization & mean/std & mean \\
HGPO length coefficient & -- & $\alpha_{\rm HGPO}=1.0$ \\
Observation mode & text-rich & text \\
Final checkpoint & step 150 & step 150 \\
Evaluation temperature & $0.4$ & $0.4$ \\
\bottomrule
\end{tabular}
\end{table}

\section{Ablation Definitions}
\label{app:baselines}
\subsection{Root-only and Full-Flat}
The online mechanism ablation uses GraphGPO with Qwen2.5-1.5B-Instruct. Root-only keeps only the first oracle target, giving the per-root term $g_r\ell_{r,0}$. Full-Flat trains the full continuation but uses one shared branch weight,
\begin{equation}
 \bar w_r=\frac{1}{L_r}\sum_{k=0}^{L_r-1}w_{r,k},\qquad
 \mathcal L_{\mathrm{flat},r}=g_r\bar w_r\sum_{k=0}^{L_r-1}\ell_{r,k}.
\label{eq:appendix_flat}
\end{equation}
SERO uses the same oracle branch but keeps the step-specific exclusive-prefix weights $w_{r,k}$. These three variants correspond directly to the ablation reported in Table~\ref{tab:ablation}.

\subsection{Fixed-Bank Controls}
The fixed-bank experiment in Appendix~\ref{app:fixed_bank} removes differences in collected oracle roots by replaying the same cached oracle branches from the same starting checkpoint. Flat, Depth decay, and SERO match the total supervision mass on every branch; their auxiliary objectives differ only in where that mass is assigned along the continuation.

\section{Limitations and Future Work}
\label{limitations}

SERO assumes that the training environment can be restored to a policy-visited state and that an executable oracle is available during training, and our experiments are limited to ALFWorld and WebShop. 
When exact restoration is unavailable, learned world models and diffusion-based generators could synthesize the observations along an oracle continuation~\citep{ha2018worldmodels,hafner2023dreamerv3,yang2024unisim,bruce2024genie,ho2020ddpm,rombach2022ldm,wang2026unireason,wang2026deepgen}. 
Extending SERO to visually grounded embodied settings~\citep{shridhar2020alfred} would require grounding the objects named in oracle commands across successive observations. Vision-language pretraining and grounding~\citep{radford2021clip,li2022glip,kirillov2023sam,yang2025moma}, multimodally guided segmentation and localization in images and videos~\citep{li2022lseg,yang2022lavt,yang2024multimodal,yang2024remamber,zhou2022avs,gao2017tall,ma2022fusioner,ma2023attrseg,zhang2023prototypical,mao2025safire,liu2023autr,liu2023annotationfree,ju2023constraintunion,wang2025contrastunity}, and diffusion-based dense prediction~\citep{xu2023odise,ma2023diffusionseg,ma2025freesegdiff,yang2026genmask} are natural tools for this, together with 3D scene representations~\citep{mildenhall2020nerf,kerbl2023gaussian,shi2022darf}. 
Finally, oracle roots are sampled uniformly from failed rollouts, so rare failure types may be under-supervised; prioritized sampling~\citep{schaul2016per} and techniques for domain shift and long-tailed data~\citep{ganin2016dann,cui2019classbalanced,ye2021uda,zhang2024rareness} may help prioritize such roots, and scaling beyond 7B models and two benchmarks is left for future work.

\end{document}